\documentclass[11pt,a4paper]{article}
\usepackage[T1]{fontenc}
\usepackage{times}
\usepackage{latexsym}
\usepackage{amsmath}
\usepackage{url}
\usepackage{multirow}
\usepackage{graphicx}
\usepackage{graphics}
\usepackage{enumitem}
\usepackage{subcaption}
\usepackage{caption}
\usepackage{amssymb}
\usepackage{verbatim}
\usepackage[normalem]{ulem} 
\usepackage{makecell}
\usepackage{array}
\usepackage{url}
\usepackage{comment}
\usepackage{multirow}
\usepackage{gensymb}
\usepackage{float}

\newcommand{\MK}[1]{\textcolor{black}{#1}}
\newcommand{\Ananya}[1]{\textcolor{black}{#1}}
\newcommand{\Akash}[1]{\textcolor{black}{#1}}

\usepackage[acceptedWithA]{tacl2018v2}
\usepackage{multicol}
\usepackage{array}
\newcolumntype{L}[1]{>{\raggedright\let\newline\\\arraybackslash\hspace{0pt}}m{#1}}
\newcolumntype{C}[1]{>{\centering\let\newline\\\arraybackslash\hspace{0pt}}m{#1}}
\newcolumntype{R}[1]{>{\raggedleft\let\newline\\\arraybackslash\hspace{0pt}}m{#1}}

\usepackage{xspace,mfirstuc,tabulary}

\definecolor{burgundy}{rgb}{0.5, 0.0, 0.13}
\definecolor{brightpink}{rgb}{1.0, 0.0, 0.5}
\definecolor{brinkpink}{rgb}{0.98, 0.38, 0.5}
\definecolor{mypink3}{cmyk}{0, 0.7808, 0.4429, 0.1412}
\definecolor{applegreen}{rgb}{0.55, 0.71, 0.0}
\definecolor{ao(english)}{rgb}{0.0, 0.5, 0.0}
\definecolor{britishracinggreen}{rgb}{0.0, 0.26, 0.15}
\definecolor{amber(sae/ece)}{rgb}{1.0, 0.49, 0.0}
\definecolor{cardinal}{rgb}{0.77, 0.12, 0.23}
\definecolor{darkorange}{rgb}{1.0, 0.55, 0.0}
\definecolor{frenchblue}{rgb}{0.0, 0.45, 0.73}
\definecolor{fulvous}{rgb}{0.86, 0.52, 0.0}
\definecolor{mahogany}{rgb}{0.75, 0.25, 0.0}
\definecolor{persimmon}{rgb}{0.93, 0.35, 0.0}
\definecolor{safetyorange(blazeorange)}{rgb}{1.0, 0.4, 0.0}
\definecolor{sapphire}{rgb}{0.03, 0.15, 0.4}
\definecolor{ultramarine}{rgb}{0.07, 0.04, 0.56}
\definecolor{ultramarineblue}{rgb}{0.25, 0.4, 0.96}
\definecolor{persianindigo}{rgb}{0.2, 0.07, 0.48}
\definecolor{persianblue}{rgb}{0.11, 0.22, 0.73}
\definecolor{palatinateblue}{rgb}{0.15, 0.23, 0.89}
\definecolor{palatinatepurple}{rgb}{0.41, 0.16, 0.38}
\definecolor{vividburgundy}{rgb}{0.62, 0.11, 0.21}
\definecolor{richmaroon}{rgb}{0.69, 0.19, 0.38}
\definecolor{springbud}{rgb}{0.65, 0.99, 0.0}
\definecolor{wildwatermelon}{rgb}{0.99, 0.42, 0.52}
\definecolor{yellow}{rgb}{1.0, 1.0, 0.0}
\definecolor{forestgreenw}{rgb}{0.13, 0.55, 0.13}
\definecolor{pastelgreen}{rgb}{0.47, 0.87, 0.47}
\definecolor{pastelorange}{rgb}{1.0, 0.7, 0.28}
\definecolor{pastelyellow}{rgb}{0.99, 0.99, 0.59}
\definecolor{pastelpink}{rgb}{1.0, 0.82, 0.86}
\definecolor{pastelviolet}{rgb}{0.8, 0.6, 0.79}
\definecolor{pastelred}{rgb}{1.0, 0.41, 0.38}
\definecolor{tomato}{rgb}{1.0, 0.39, 0.28}
\definecolor{palerobineggblue}{rgb}{0.59, 0.87, 0.82}
\definecolor{wisteria}{rgb}{0.79, 0.63, 0.86}
\definecolor{violetw}{rgb}{0.93, 0.51, 0.93}

\newif\iftaclinstructions
\taclinstructionsfalse 
\iftaclinstructions
\renewcommand{\confidential}{}
\renewcommand{\anonsubtext}{(No author info supplied here, for consistency with
TACL-submission anonymization requirements)}
\newcommand{\instr}
\fi

\iftaclpubformat 

\else

\fi

\title{Improving Dialog Evaluation with a Multi-reference Adversarial Dataset and Large Scale Pretraining}

\author{Ananya B. Sai\Thanks{The first two authors worked equally towards the project.} \and Akash Kumar Mohankumar\footnotemark[1] \and Siddhartha Arora \and Mitesh M. Khapra \\
\{ananya, miteshk\}@cse.iitm.ac.in , \{makashkumar99, sidarora1990\}@gmail.com \\
Robert-Bosch Centre for Data Science and Artificial Intelligence\\ Indian Institute of Technology, Madras}
\date{}

\begin{document}
\maketitle
\begin{abstract}
There is an increasing focus on model-based dialog evaluation metrics such as ADEM, RUBER, and the more recent BERT-based metrics. These models aim to assign a high score to \textit{all} relevant responses and a low score to all irrelevant responses. Ideally, such models should be trained using multiple relevant and irrelevant responses for any given context. However, no such data is publicly available, and hence existing models are usually trained using a \textit{single} relevant response and multiple \textit{randomly selected} responses from other contexts (random negatives). To allow for better training and 
robust evaluation of model-based metrics, we introduce the \textit{DailyDialog++} dataset, consisting of (i) five relevant responses for each context and (ii) five \textit{adversarially crafted} irrelevant responses for each context. Using this dataset, we first show that even in the presence of multiple correct references, $n$-gram based metrics and embedding based metrics do not perform well at separating relevant responses from even \textit{random} negatives. While model-based metrics perform better than $n$-gram and embedding based metrics on random negatives, their performance drops 
substantially 
when evaluated on adversarial examples.
To check if large scale pretraining could help, we propose a new BERT-based evaluation metric called DEB, which is pretrained on 727M Reddit conversations and then finetuned on 
our dataset. DEB significantly outperforms existing models, showing better correlation with human judgements and better performance on random negatives (88.27\% accuracy). However, its performance again drops 
substantially, when evaluated on adversarial responses, thereby highlighting that even large-scale pretrained evaluation models are not robust to the adversarial examples in our dataset. 
The dataset\footnote{Dataset: https://iitmnlp.github.io/DailyDialog-plusplus/} and code\footnote{Code: https://github.com/iitmnlp/Dialogue-Evaluation-with-BERT} are publicly available.
\end{abstract}

\section{Introduction}
    Open-domain conversational systems are increasingly in demand for several applications ranging from personal digital assistants to entertainers for recreation. While several automated dialogue agents such as Siri, Alexa, Cortana and Google Assistant have been built and deployed, there is no good automatic evaluation metric to measure the quality of their conversations. Researchers have usually adopted n-gram based metrics \cite{bleu, meteor, rouge} or embedding based metrics \cite{vectorextrema,greedymatch,bertscore} to compare the model's response with a \textit{single} reference. These metrics assume that a valid response should be semantically or lexically similar to the reference without taking the context of the conversation into consideration. However, in open domain conversations, a given context can have a wide range of possible responses that may be lexically and semantically very different from each other.  For example, the context, ``I like dancing and swimming, what about you?'' can be responded to with ``I paint in my free time'' or ``I do not have time for hobbies right now'', both of which are valid responses. As a result, $n$-gram and word embedding based metrics, which rely on lexical and/or semantic match, correlate very weakly with human judgements for dialogue evaluation  \cite{DBLP:conf/emnlp/LiuLSNCP16HowNot}. 
    
    Given the shortcomings of context-agnostic $n$-gram and embedding based metrics, the focus has now shifted to building neural network based, trainable dialogue evaluation models \cite{DBLP:conf/acl/LoweNSABP17ADEM, DBLP:conf/aaai/TaoMZY18ruber, shimanaka19, ghazarian19}. Such models are trained to identify whether a given response can be considered as a valid continuation of the given context or not. In other words, the model should (i) assign a high score to \textit{all} relevant responses no matter how diverse they are and (ii) assign a low score to all irrelevant responses, preferably with a clear margin of separation from relevant responses. Although there exist several open-domain dialogue datasets \cite{DBLP:conf/semco/ForsythandM07NPSChat, DBLP:conf/lrec/Tiedemann12OpenSub, DBLP:conf/naacl/RitterCD10Twit, DBLP:conf/ijcnlp/LiSSLCN17DailyOri} that are used for training dialogue response generation systems, they are not suitable for training and testing such evaluation models. This is because these datasets have only \textit{a single} relevant response and no irrelevant responses. Irrelevant responses can of course be generated by sampling random utterances from other contexts, but such examples typically do not have any overlap with the context and hence are easier for the model to distinguish from relevant responses (as we will show in our results later). We refer to the randomly sampled responses as \emph{random} negatives.
    
    Some efforts have been made to build dialog datasets with multiple relevant responses (\textit{i.e.}, multiple references), but these datasets are either very small (1000 contexts) \cite{DBLP:conf/emnlp/MogheABK18Holle,DBLP:journals/corr/abs-1907-10568InvestEval} or automatically constructed from Reddit conversations, hence, potentially noisy  \cite{DBLP:conf/naacl/GaoLZBGGD19redditmultiforgenerationdiversity}. Further, these datasets do not have any carefully crafted \emph{adversarial} irrelevant responses. We define an \emph{adversarial} irrelevant response as one which has a significant word overlap with the context but is still an irrelevant response (hence harder to identify than randomly selected irrelevant examples, which may not have any relation to the context). To overcome this limitation of existing datasets, we propose a large scale multi-reference dataset, \textit{DailyDialog++}, which is an extension of the DailyDialog dataset. In particular, for each of the 19K contexts derived from DailyDialog, we collect additional 5 reference responses with the help of human annotators. Further, for $\sim$11K contexts in DailyDialog, we also ask human annotators to carefully craft irrelevant responses which have a significant word overlap with the context. This dataset will be made publicly available and help towards better training and more robust evaluation of \Akash{dialogue evaluation metrics}.

    Using this dataset, we extensively evaluate a wide range of $n$-gram-based and embedding-based metrics. In particular, we compute (i) the correlation of these metrics with binary human judgements and (ii) the accuracy obtained by using the scores assigned by the metrics to classify relevant/irrelevant responses. The performance of these metrics improves when presented with multiple references as opposed to a single reference, but they still leave a lot to be desired. On the other hand, most model-based evaluation metrics, when trained and evaluated using multiple relevant and random negative responses, perform significantly better than the $n$-gram-based and embedding-based methods. However, their performance drops substantially on the adversarial examples in our dataset. 
    
    Lastly, one could argue that dialog evaluation metrics could be improved by pretraining on large amounts of data. To check if this is indeed the case, we propose a new BERT-based evaluation metric called DEB (Dialog Evaluation using BERT), which is pretrained on 727M Reddit conversations. Indeed, this model performs significantly better on random negatives with an accuracy of 88.27\% in distinguishing the positive and random negative responses. It also correlates well with human judgments on responses generated by five dialog generation systems \cite{HRED, VHRED, VHCR, zhang2019dialogpt}. In particular, the Spearman rank correlation between human scores and DEB scores is 0.52 at the response level scores and 0.70 at the system level scores, calculated by aggregating the scores on all responses by each system. However, once again, when evaluated on adversarial examples from our dataset, its performance drops substantially, underscoring that even large-scale pretrained models are not robust to adversarial examples. 

\section{Proposed Dataset}
\begin{table*}[t]
    \centering \small
    \begin{tabular}{L{4cm}|L{4cm}|L{7cm}}
         Context & Valid responses & Invalid, adversarial responses \\
         \hline
         \textcolor{red}{FS:} Can you do push-ups ? \newline \textcolor{blue}{SS:}  Of course I can . It's a piece of cake ! Believe it or not , I can do 30 push-ups a minute. \newline \textcolor{red}{FS:}  Really ? I think that's impossible ! \newline \textcolor{blue}{SS:}  You mean 30 push-ups ? \newline \textcolor{red}{FS:}  Yeah ! & SS: You don't believe me, do you? \newline SS: Start your timer, here we go. \newline SS: Watch me do it.\newline SS: That's because you can't do it.\newline SS: You don't know that I am a fitness trainer, do you ? & SS: \textcolor{purple}{\underline{Push up}} the window and look out for a \textcolor{purple}{\underline{minute}}\newline SS: Would you like to eat a \textcolor{purple}{\underline{piece of cake}} before \textcolor{purple}{\underline{gym}}?\newline SS: I like watching the Ripley's \textcolor{purple}{\underline{Believe it or Not}} show where they discuss nearly \textcolor{purple}{\underline{impossible}} \textcolor{purple}{\underline{feats}} and \textcolor{purple}{\underline{gymnastics}}\newline SS: I have enough \textcolor{purple}{\underline{time}} for my \textcolor{purple}{\underline{treadmill exercises}}\newline SS: Are you asking me to do 40 \textcolor{purple}{\underline{squats}}?\\
         \hline
    \end{tabular}
    \caption{Examples from DailyDialog++ dataset with the context consisting of 2 speakers [annotated as FS (First Speaker) and SS (Second Speaker)], and multiple reference responses and adversarial negative responses. The underlined, purple colored words in the adversarial responses are those that overlap or are closely related to the theme or words in the context}
    \label{tab:dataset_egs}
\end{table*}
    Our goal was to build a dataset with manually-created multiple relevant and \Akash{adversarial} irrelevant responses. For this, we wanted to start with an existing base dataset, which already has one relevant response for every context, and then extend it to include multiple responses. For the base dataset, we considered several popular datasets such as Twitter \cite{DBLP:conf/naacl/RitterCD10Twit}, Reddit \cite{reddit}, Open Subtitles \cite{DBLP:conf/lrec/Tiedemann12OpenSub}, NPS Chat \cite{DBLP:conf/semco/ForsythandM07NPSChat}, PersonaChat \cite{personachat} and DailyDialog \cite{DBLP:conf/ijcnlp/LiSSLCN17DailyOri}. Of these, Twitter and Reddit are generally considered noisy, so we chose not to use either of them as the base dataset. Similarly, Open Subtitles and NPS Chat did not have speaker aligned utterances and hence were not suitable for our purposes. We found that the DailyDiaog dataset was clean, human-written, readily available, and covered a diverse set of generic topics such as \textit{ordinary life, school life, tourism, attitude \& emotion, relationship, health, work, politics, culture \& education} and \textit{finance}. It contains a total of 13K conversations with an average of 8 turns between exactly 2 speakers. Alternatively, we could have also chosen PersonaChat, which is of a similar size and also contains chit-chat style conversations, but we chose the antecedent DailyDialog dataset.

For shorter conversations in DailyDialog (having less than 8 turns) we collected multiple relevant responses only for the last utterance. For longer conversations (having 8 turns or more), we divided the conversation into two or more smaller chunks and collected multiple relevant responses for the last utterance in every chunk. In this way, from the 13K conversations\footnote{Out of the 13K conversations released in DailyDialog, we found that a good number of contexts were repeated, either with slightly different spellings or through some subtle differences such as representing numbers using digits versus using words. We filtered out the repetitions and worked with the remaining $\sim$11K contexts.} in DailyDialog, we were able to create 19K sub-conversations with multiple relevant responses for the last utterance in each sub-conversation or context. The responses were created by in-house annotators. Each context was shown to 2-3 annotators, and each of them was asked to generate 1-3 alternative responses for the last utterance, capping the total number of alternative responses to 5 (in addition to the one response already available in DailyDialog). The annotators were strictly instructed to avoid short generic responses (``Okay'', ``Thank you'', ``Sure'', \textit{etc.}), and write longer meaningful responses containing at least 8-10 words. These responses were then verified (and if needed, corrected and re-validated) by a different set of annotators.

\subsection{Adversarial irrelevant responses}

In addition to collecting multiple relevant responses for each context, we also wanted to collect irrelevant responses for each context. 
Most of the models which are trained for the task of dialogue evaluation (and dialogue generation) \cite{DBLP:conf/aaai/TaoMZY18ruber, ghazarian19, li2017adversarial} procure irrelevant responses by randomly sampling responses from other contexts.
Such random negatives are often entirely out of context (unrelated) and hence are too easy for the model to distinguish. To allow for a more critical or adversarial examination of dialogue evaluation systems, we propose creating adversarially crafted irrelevant responses that have lexical or semantic overlap with the context but are still unacceptable as valid responses.

For obtaining such tricky negative responses, the annotators were asked to choose some words from the context and use them directly or indirectly while writing the responses. Indirect usage here refers to using words closely related to the context words.  For example, using synonyms, antonyms, homonyms, subwords, or other words that are known to frequently co-occur with the words in the context (e.g., the words ``flexibility'' and ``injuries'' co-occur with ``acrobatics''). Once again, each context was shown to 2-3 annotators, and each of them was asked to generate 1-3 adversarially crafted responses for the last utterance, capping the total number of alternative responses to 5. Each response was then validated by two different annotators. The validating annotators were instructed to either eliminate or modify the responses that were not negative or were borderline. A final check was made by one more evaluator to ensure that the responses were adversarially crafted, irrelevant, and grammatically correct. We collected 5 such responses for 11429 contexts.  Table \ref{tab:dataset_egs} shows examples of relevant and irrelevant responses in our dataset and Table \ref{tab:dataset_stats} shows some statistics about our dataset.
\begin{table}[h]
    \centering
    \resizebox{0.8\linewidth}{!}{
    \begin{tabular}{l|r}
    \hline
        Total \# of contexts & 19071 \\
        Avg. \# of turns per context & 3.31 \\
        Avg. \# of words per context &  45.32\\
        Avg. \# of words per utterance & 13.55 \\\hline
        \# of contexts with 5 relevant responses &  19071\\
        \# of contexts with 5 adv. irrelevant responses  & 11429 \\
        Avg. \# of words per relevant response & 10.13 \\
        Avg. \# of words per irrelevant response & 13.8 \\\hline
    \end{tabular}}
    \caption{DailyDialog++ Dataset Statistics}
    \label{tab:dataset_stats}
\end{table}

\MK{We acknowledge that, in practice, a given context can have a large number of relevant responses ($>> 5$). However, exhaustively collecting all such responses is prohibitively expensive and time consuming. While it is desired to have even more than 5 responses for every context, we believe that having at least 5 is a good starting point given the dearth of such multi-reference conversation datasets. The proposed dataset thus serves as a pragmatic substitute for an ideal dataset which would have contained a large number of responses per context. Having said that, we would also like to point out that the value of the proposed dataset goes beyond having multiple relevant references as it is also the first dataset containing adversarial irrelevant responses for given contexts. }
\section{Existing metrics \label{sec:existing-metrics}}
In this section, we present a brief overview of the existing automatic metrics used for dialogue evaluation. The existing metrics can be broadly classified into two categories, \textit{viz.}\ (i) Untrained metrics, and (ii) Trained metrics. Untrained evaluation metrics, usually adopted from the NLG literature, use a predefined formula to compare the candidate response with a reference without taking the context into account. On the other hand, trained metrics are  \Akash{usually} trained specifically for the task of dialogue response evaluation to identify valid and invalid responses for a given context. 

\subsection{Untrained Metrics}
Untrained metrics can be further sub-classified into (i) n-gram based, (ii) word embedding based, and (iii) contextualized embedding based metrics.  

\textbf{N-gram based:} 
\noindent N-gram based metrics score a candidate response based on the amount of n-gram overlap it has with a given reference. BLEU \cite{bleu}, ROUGE-L \cite{rouge} and METEOR \cite{meteor} are among the most commonly adopted n-gram based metrics to evaluate dialogue systems. BLEU is calculated using n-gram precision scores between the candidate response and the reference. ROUGE-L \cite{rouge} is based on the F-measure of the longest common subsequence between the candidate and reference responses. METEOR \cite{meteor} relaxes the exact match criteria by including word stems, synonyms, and paraphrases. 
\Ananya{More recently, \newcite{deltaBLEU} proposed deltaBLEU which takes in multiple references and rewards n-gram matches with positive references and penalizes the matches with the negative references.}

 \textbf{Word embedding based:} These methods use word embeddings to compute the similarity between the candidate response and the reference response. The most commonly used word embedding based metrics are Embedding Average \cite{DBLP:journals/corr/WietingBGL15a}, Vector Extrema \cite{vectorextrema} and Greedy Matching \cite{greedymatch}. Embedding Average defines a sentence embedding as the average word embedding of the constituent words. The final score is calculated using the cosine similarity of candidate and reference sentence embeddings. Vector Extrema \cite{vectorextrema} instead computes the sentence embedding by taking the most extreme value for each dimension. In other words, the value of the $i$-th dimension of the sentence embedding is computed by taking a maximum over the $i$-th dimension of all words in the sentence. Greedy Matching \cite{greedymatch} first computes the maximum cosine similarity that every word in the candidate response has with any word in the reference response. 
 Similarly, the highest cosine similarity for each of the reference words with any of the candidate response words is calculated. 
 The similarity between the candidate response and reference response is then computed by taking an average of the maximum cosine similarities computed above.

 \textbf{BERTScore:} Recently, \newcite{bertscore} proposed BERTScore, which uses contextualized word embeddings of the candidate and reference sentences to compute the score. BERTScore is similar to greedy matching but uses contextualized embeddings from BERT instead of static word embeddings. 
\subsection{Trained Metrics}
\textbf{ADEM:} Automatic Dialogue Evaluation Model (ADEM) \cite{DBLP:conf/acl/LoweNSABP17ADEM} uses pretrained vector representations of the the dialogue context $\mathbf{c}$, reference response $\mathbf{r}$, and proposed response $\mathbf{\hat{r}}$ to compute the evaluation score as follows:
\begin{equation}
    \mbox{Score}(\mathbf{c}, \mathbf{r},\mathbf{\hat{r}}) = (\mathbf{c}^T \mathbf{M} \mathbf{\hat{r}} + \mathbf{r}^T \mathbf{N} \mathbf{\hat{r}} - \alpha)/ \beta
\end{equation}
where $\mathbf{M}$, $\mathbf{N} \in \mathbb{R}^{n \times n}$ are learned matrices, and $\alpha, \beta$ are scalar constants used to
re-scale scores in the range $[1, 5]$. The context, proposed response and reference response are encoded using a Hierarchical RNN (H-RNN) encoder consisting of utterance-level and context-level RNNs. The H-RNN encoder is pretrained on a Twitter dataset \cite{twitterdataset} in a generative setup using the latent variable hierarchical recurrent encoder decoder (VHRED) model \cite{VHRED}. The weight matrices, $\mathbf{M}, \mathbf{N}$, are later finetuned for the task of dialogue response evaluation.

\noindent \textbf{RUBER:} \cite{DBLP:conf/aaai/TaoMZY18ruber} introduced an unreferenced evaluation model consisting of GRU encoders \cite{gru} to measure the relatedness between the dialogue context and a given response. The authors train the model on Chinese dialogue data with the hinge loss objective.

\noindent \textbf{BERT regressor\footnote{Since we couldn't find an exact name for the evaluator model by \newcite{shimanaka19} , we adopt the name, `BERT regressor' from their paper's title.}:} \newcite{shimanaka19} propose a BERT based evaluation model to score a candidate sentence based on a reference. Unlike BERTScore, the BERT model is finetuned to predict human judgement scores from the concatenated reference and candidate sentence. 

\noindent \textbf{BERT+DNN\footnote{Due to the lack of a specific name for the models in \newcite{ghazarian19}, we refer to the model adopted from their work as `BERT+DNN'}:} \newcite{ghazarian19} use contextualized embeddings to compute a relatedness score between the dialogue context and response. The best performing model of \newcite{ghazarian19} consists of a multi-layer perceptron that takes the concatenation of contextualized representations of the context and response as input. The contextualized representations are obtained by max-pooling the respective BERT embeddings for each token. Note that the BERT embeddings are not finetuned.

\section{Dialogue Evaluation using BERT \label{sec:DEB} }

In the last two years, a lot of success in NLP has been driven by large pretrained transformer-based models \cite{gpt2, DBLP:conf/naacl/DevlinCLT19bert, ERNIE}. These models are typically trained with a language model objective and leverage large amounts of unlabeled data. However, none of the trained metrics discussed in the previous section leverage pretraining on large-scale dialogue corpora. With the hope that such pretraining should help dialog evaluation models also, we introduce DEB (Dialog Evaluation using BERT) which is trained using a masked language model objective (similar to BERT) and a modified next response prediction objective. 

We set up the the task of next response prediction as one of identifying whether the given response is a valid next response for the given context. Formally, given a context $\mathbf{C} = \{w^c_1, \dots, w^c_n\}$ and a response $\mathbf{R} = \{w^r_1, \dots, w^r_m\}$, we first pass the concatenated sequence $\ \mathbf{U} =$ $\{\mbox{[CLS]}, w^c_1,$ $\dots, w^c_n, \mbox{[SEP]}, w^r_1,$ $\dots, w^r_m \}$ through the BERT transformer and obtain $\mathbf{H}_{cls} \in \mathbb{R}^{H}$, the last-layer activations corresponding to the special [CLS] token. We then make our final next response predictions as follows: $\hat{y} = \mbox{softmax}(\mathbf{W} \mathbf{H}_{cls})$, where $ \mathbf{W} \in \mathbb{R}^{2 \times H}$ is a learnable matrix. We use cross entropy loss with binary targets for the next-response prediction. In addition, we use the standard masked language model objective by randomly masking 15\% of the words in $\mathbf{C}$ and $\mathbf{R}$. 

Note that the proposed model is a straightforward extension of the standard BERT model used for language modeling. We do not claim any novelty on this front. The key contribution here is to assess if pretraining on large-scale dialogue corpora improves the performance of dialogue evaluation metrics. Existing BERT-based evaluation metrics \cite{shimanaka19,ghazarian19} do not use such pretraining on any large-scale, domain-related corpora. 
In other words, they do not leverage the more successful recipe of (i) pretraining with a masked language modeling objective and (ii) finetuning with a task-specific objective (dialog evaluation in this case). The idea behind DEB is to check if this successful recipe can be replicated for dialog evaluation, making use of the dialogues in the large-scale Reddit corpus.

\subsection{Training details} For pretraining, we use a massive open-domain dialogue dataset of Reddit comments from 2005 to 2019 consisting of 256M threads with a total of 3.68B comments. From this dataset, we extracted a total of 727M \{\textit{context}, \textit{positive response}\} pairs with 654M for training and 73M for testing following the method described in \newcite{reddit}. We used an equal number of negative responses by randomly sampling responses from other contexts. We use the BERT base model with 110M parameters consisting of 12 layers, 768 dimensional hidden space, and 12 attention heads per layer in all our experiments. 
We finetune the pretrained DEB model on our DailyDialog++ dataset for 1 epoch (we did not see any advantage of finetuning beyond 1 epoch). Note that during finetuning we only use the next response prediction objective.

\section{Experimental Setup}
\MK{Our goal is to check if the adversarial responses in our dataset, which are specifically crafted to target context-dependent model-based metrics (such as ADEM, RUBER, BERT+DNN, and DEB), indeed affect the performance of such models. To do so,} 
we first need to benchmark the models' performance on random negatives and then check if the performance drops when evaluated on adversarial examples. 
Hence, in this section, we describe (i) the process of creating and validating such random negatives and (ii) the process used for training model-based metrics.

We randomly divide our dataset into train (80\% contexts), validation (10\% contexts) and test (10\% contexts) splits. Note that, adversarial negatives are not used for training or finetuning the models unless explicitly specified. 

\subsection{Creating \& validating random negatives} 
 For every context in our dataset, which has 5 relevant responses, we also sample 5 random negatives. While sampling random negatives, we avoid short responses that may be generic and relevant for any context. To verify whether the sampled random negatives were indeed irrelevant, we asked human annotators to manually check 500 such sampled responses. More specifically, we showed them the original context and the sampled random negative response and asked them if it was a relevant or irrelevant response. In 95\% of the cases, the annotators confirmed that the random negative response was irrelevant, thereby confirming that a random sampling strategy indeed results in irrelevant responses (although they may not be as hard as our adversarial negative examples as shown later). 


\subsection{Pretraining \& finetuning trained metrics} 
We describe the pretraining and finetuning procedure for the various models used in our analysis below.

\noindent\textbf{ADEM:} As previously mentioned in Section \ref{sec:existing-metrics}, ADEM was pretrained on Twitter corpus using the VHRED setup and then finetuned for dialogue response evaluation. We take this publicly available model and finetune it further using our DailyDialog++ dataset with a target of 5 for positive responses and 1 for random negatives. The reference response could be any of the other four relevant responses. Note that ADEM produces a score on a scale of 1 to 5 whereas the other models produce a score on a scale of 0 to 1. For easier comparison, we scale the output of ADEM so that it lies in the range of 0 to 1. 

\noindent \textbf{BERT regressor:} We finetune the publicly available pretrained BERT base model (110M parameters) on our DailyDialog++ dataset. We train the model with a label of 1 for positive responses and 0 for random negative responses using any one of the other four positive responses as the reference. We train the model using cross-entropy loss and follow the same set of hyper-parameters as used by \newcite{shimanaka19} during finetuning. 

\noindent \textbf{BERT+DNN}: We use the best performing model from \newcite{ghazarian19}, which consists of a three layered feed-forward neural network 
and uses pretrained BERT embeddings as input. We train the model on our DailyDialog++ dataset with random negatives using cross entropy loss. 

\noindent\textbf{RUBER and RUBER-Large:} We experiment with two variants of \newcite{DBLP:conf/aaai/TaoMZY18ruber}'s models with different sizes, \textit{viz,} (i) RUBER (34M parameters), which consists of single-layer GRUs with a hidden size of 1024, and (ii) RUBER-Large (236M parameters), which consists of two layered GRUs with a hidden size of 2048. As shown in \newcite{vaswanietal2017}, the training time for RNN based architectures is very high when compared to the transformer models that allow much greater parallelization. We observed an estimated time of over 200 days to train the RUBER-Large model on the 727M Reddit corpus on a 1080ti GPU, thereby making it practically infeasible to train such models on large-scale datasets. Taking the computational costs into consideration, we pretrained RUBER and RUBER-Large on a sample of 20M contexts with relevant and random irrelevant responses from Reddit. We then finetuned these models on our proposed dataset with random negatives. \footnote{We agree that this may not be a fair comparison but we we were constrained by the inherent limitations of such RNN-based, sequential models which make large-scale pretraining prohibitively expensive and time consuming.}

\Akash{\noindent \textbf{DEB:} We pretrained DEB on the entire 727M Reddit corpus using the masked language model and the modified next response prediction objective. Pretraining DEB took 4 days on a single Google Cloud TPUv2. We achieved a test accuracy of 90\% on the next response prediction task and a perplexity of $15.47$ (58\% accuracy) on the masked language modelling task in the pretraining corpus. We then finetuned DEB on our dataset with random negatives. }

\setlength{\tabcolsep}{16pt}
\renewcommand{\arraystretch}{1.0}
\begin{table*}[]
\centering
\resizebox{0.90\textwidth}{!}{
\begin{tabular}{l|c|c|c|c|c|c|c|c}
\multicolumn{1}{c|}{\multirow{3}{*}{Metric}} & \multicolumn{4}{c|}{Point Biserial Correlation (p-value)}                                                                           & \multicolumn{4}{c}{Accuracy in percentage}                                                                      \\ \cline{2-9} \cline{2-9} \cline{2-9} \cline{2-9} 
\multicolumn{1}{c|}{}                        & \multicolumn{1}{c|}{\multirow{2}{*}{Single}} & \multicolumn{3}{c|}{Multiple}                       & \multicolumn{1}{c|}{\multirow{2}{*}{Single}} & \multicolumn{3}{c}{Multiple}                       \\ \cline{3-5} \cline{7-9} 
\multicolumn{1}{c|}{}                        & \multicolumn{1}{l|}{}                        & \multicolumn{1}{c|}{Avg} & \multicolumn{1}{c|}{Max} & \multicolumn{1}{c|}{Standard} & \multicolumn{1}{l|}{}                        & \multicolumn{1}{c|}{Avg} & \multicolumn{1}{c|}{Max} & \multicolumn{1}{c}{Standard} \\ \Xhline{3\arrayrulewidth}
BLEU-1 & 0.26  \fontsize{9}{9}\selectfont{(<1e-9)} & 0.42 \fontsize{9}{9}\selectfont{(<1e-9)} &   0.41 \fontsize{9}{9}\selectfont{(<1e-9)} & 0.41 \fontsize{9}{9}\selectfont{(<1e-9)} & 61.26 & 68.60 & 68.75 & 70.36\\ 
BLEU-2 & 0.22 \fontsize{9}{9}\selectfont{(<1e-9)} & 0.39 \fontsize{9}{9}\selectfont{(<1e-9)} & 0.36 \fontsize{9}{9}\selectfont{(<1e-9)} & 0.40 \fontsize{9}{9}\selectfont{(<1e-9)} & 58.09 & 68.26 & 68.37 & 68.66\\ 
BLEU-3 & 0.14 \fontsize{9}{9}\selectfont{(<1e-9)} & 0.26 \fontsize{9}{9}\selectfont{(<1e-9)} & 0.24 \fontsize{9}{9}\selectfont{(<1e-9)} & 0.28 \fontsize{9}{9}\selectfont{(<1e-9)} & 53.11 & 58.85 & 58.90 & 58.89\\ 
BLEU-4 & 0.08 \fontsize{9}{9}\selectfont{(<1e-9)} & 0.17 \fontsize{9}{9}\selectfont{(<1e-9)} & 0.15 \fontsize{9}{9}\selectfont{(<1e-9)} & 0.18 \fontsize{9}{9}\selectfont{(<1e-9)} & 51.16 & 53.56 & 53.56 & 53.50 \\ 
METEOR & 0.23 \fontsize{9}{9}\selectfont{(<1e-9)} & 0.40 \fontsize{9}{9}\selectfont{(<1e-9)} & 0.41 \fontsize{9}{9}\selectfont{(<1e-9)} & - & 59.77 & 68.51 & 68.01 & - \\ 
ROUGE-L & 0.23 \fontsize{9}{9}\selectfont{(<1e-9)} & 0.41 \fontsize{9}{9}\selectfont{(<1e-9)} & 0.40 \fontsize{9}{9}\selectfont{(<1e-9)} & 0.37 \fontsize{9}{9}\selectfont{(<1e-9)} & 59.47 & 67.89 & 68.25 & 68.43\\  
deltaBLEU \cite{deltaBLEU} & - & - & - & 0.29 \fontsize{9}{9}\selectfont{(<1e-9)} & - & - & - & 64.89 \\ 
\hline
Embed Avg & 0.23 \fontsize{9}{9}\selectfont{(<1e-9)} & 0.25 \fontsize{9}{9}\selectfont{(<1e-9)} & 0.23 \fontsize{9}{9}\selectfont{(<1e-9)} & - & 61.27 & 61.56 & 62.67 & - \\ 
Vec Extr \cite{vectorextrema} & 0.24 \fontsize{9}{9}\selectfont{(<1e-9)} & 0.35 \fontsize{9}{9}\selectfont{(<1e-9)} & 0.33 \fontsize{9}{9}\selectfont{(<1e-9)} & - & 59.22 & 63.70 & 63.90 & -\\
GreedyMatch \cite{greedymatch} & 0.24 \fontsize{9}{9}\selectfont{(<1e-9)} & 0.36 \fontsize{9}{9}\selectfont{(<1e-9)} & 0.32 \fontsize{9}{9}\selectfont{(<1e-9)} & - & 60.02 & 63.99 & 65.56 & - \\ 
BERTScore \cite{bertscore} & 0.29 \fontsize{9}{9}\selectfont{(<1e-9)} & 0.39 \fontsize{9}{9}\selectfont{(<1e-9)} & 0.39 \fontsize{9}{9}\selectfont{(<1e-9)} & - & 63.71 & 69.05 & 68.59 & -\\ \hline
\Xhline{3\arrayrulewidth}
ADEM \cite{DBLP:conf/acl/LoweNSABP17ADEM}                  & \multicolumn{4}{c|}{0.40 \fontsize{9}{9}\selectfont{(<1e-9)}}                         & \multicolumn{4}{c}{64.74}                          \\
BERT regressor \cite{shimanaka19}                 & \multicolumn{4}{c|}{0.52 \fontsize{9}{9}\selectfont{\fontsize{9}{9}\selectfont{(<1e-9)}}}                         & \multicolumn{4}{c}{73.40}                          \\
BERT+DNN \cite{ghazarian19}                 & \multicolumn{4}{c|}{0.57 \fontsize{9}{9}\selectfont{(<1e-9)}}                         & \multicolumn{4}{c}{74.67}                          \\
RUBER \cite{DBLP:conf/aaai/TaoMZY18ruber}                 & \multicolumn{4}{c|}{0.64 \fontsize{9}{9}\selectfont{(<1e-9)}}                         & \multicolumn{4}{c}{78.18}                          \\ 
RUBER-Large \cite{DBLP:conf/aaai/TaoMZY18ruber}                 & \multicolumn{4}{c|}{0.69 \fontsize{9}{9}\selectfont{(<1e-9)}}                         & \multicolumn{4}{c}{82.36}                          \\ \Xhline{3\arrayrulewidth}
DEB (ours)           & \multicolumn{4}{c|}{\textbf{0.79*} \fontsize{9}{9}\selectfont{(<1e-9)}}                         & \multicolumn{4}{c}{\textbf{88.27*}} \\               
\end{tabular}}
\caption{Automatic evaluation metrics performance on random negatives (PBC refers to point-biserial correlation. Column subheading `Single' refers to experiments using single reference response and `Avg' and `Max' are the average and maximum aggregation strategies when using multiple reference responses. `Standard' is applicable when the metric aggregates multiple references differently. * indicates statistical significance in performance over all other metrics (with p-values <1e-9) on William’s test for comparing correlations and Chi-squared test for accuracies. p-values for individual correlations are in parenthesis}
\label{table:eval_random}
\end{table*}
\subsection{Untrained metrics with multiple references}
Untrained metrics like METEOR, Greedy Matching, etc usually work with a single reference response but can also be adapted to work with multiple reference responses. For example, for a given candidate response $c$ and a set of reference responses $r_1, r_2, r_3, ..., r_k$, we can compute the multi-reference METEOR score as:
\begin{equation*}
    METEOR_{multi} = max_{i=1}^{k} METEOR(c, r_i)
\end{equation*}
Instead of the \textit{max} function we can also use the average function. We use a similar formula for all the untrained metrics.  

\Akash{A few metrics like BLEU, deltaBLEU, and ROUGE-L have their own standard formula to incorporate multiple references.} \Ananya{BLEU calculates the number of matches for each n-gram based on the maximum number of times the n-gram occurs in common with any one of the references.} \Akash{deltaBLEU further extends the same idea to incorporate a score for each reference. We follow the implementation from \newcite{deltaBLEU} to compute the deltaBLEU scores.} \Ananya{For ROUGE-L, we follow the strategy in \newcite{sharma2017nlgeval} where the score is an F-measure of the maximum precision and maximum recall over all the references}. In addition to the average and maximum aggregations, we also report these standard multi-reference scores for BLEU, deltaBLEU and ROUGE-L. 
\section{Results}
In this section, we compare the performance of different dialog evaluation metrics in separating relevant references from (i) random negatives (ii) synthetically crafted adversarial irrelevant responses (explained below) and (iii) manually crafted adversarial irrelevant responses (as in our DailyDialog++ dataset).

\subsection{Performance on random negatives}
For every context in our test split, we obtain the scores assigned by a given metric to the 5 positive and 5 random negative responses. \Akash{In particular, we treat each of the 5 relevant and 5 random irrelevant responses as a candidate response. For all untrained metrics other than deltaBLEU, we consider the remaining 4 relevant responses as reference responses. For deltaBLEU, we consider the remaining 4 relevant responses as references with a score of 1 and the remaining 4 irrelevant responses as references with a score of -1.} We expect a good evaluation metric to provide high scores on relevant responses and low scores on the irrelevant responses. 
We then quantify the performance of all metrics using two measures. First, we compute the Point Biserial correlation (PBC) between the scores assigned by a metric and the binary target \textit{i.e.}, a score of 1 for positive responses and 0 for random negative responses.\footnote{Note that it can be shown that PBC is equivalent to the Pearson correlation when one of the variables is binary, as is the case above.} Second, we compute the classification accuracy of the metric by using a threshold and marking all responses having a score above this threshold as positive and others as negative. We use a threshold of 0.5 for the trained metrics. For all the untrained metrics, we perform a search from 0 to 1 with step size of 0.01 and select the threshold that minimizes the error rate on the validation set.\footnote{With this approach of setting a threshold, we want to be lenient with the untrained metrics and investigate how best they can be adopted. One might also think of using the median of all the scores assigned by a metric as its threshold, however, such an approach is error-prone and has several boundary conditions that would fail the purpose. We hence estimate the threshold by minimizing the risk.} Later in Section \ref{sec:box_plots_scores}, we shall observe that if we use 0.5 as the threshold, the performance of most untrained metrics would be abysmally poor. Note that for the trained metrics we found that the scores were spread evenly in the range of 0 to 1 and there was no benefit of doing a grid search to find the threshold -- a threshold of 0.5 was adequate.

In Table \ref{table:eval_random}, we report PBC and accuracy of the different untrained metrics with both single and multiple references, and the trained metrics. When evaluating using single references, we use any one of the 5 relevant responses as a reference response (other than the one being used as a candidate). We observe that with a single reference, all the untrained metrics are poor at distinguishing between the positive and random negative responses as inferred from the low accuracy and correlation values. When we use multiple responses, we observe a relatively better performance. We notice that the performance is largely similar across the aggregation techniques -- average, maximum and standard (when applicable). Metrics such as BLEU-1, METEOR, ROUGE-L and BERTScore with multiple references are able to achieve modest correlations with the binary target. Interestingly, we observe that all the word embedding based methods even in the presence of multiple references perform badly in scoring the positive and random negative responses. In contrast, trained metrics such as BERT regressor, RUBER, BERT+DNN, and DEB perform substantially better than the untrained metrics. Our proposed DEB model achieves state-of-the-art performance with an accuracy of 88.27\% and a strong correlation of 0.79. 



\begin{figure*}
    \centering
    \includegraphics[width=0.95\textwidth]{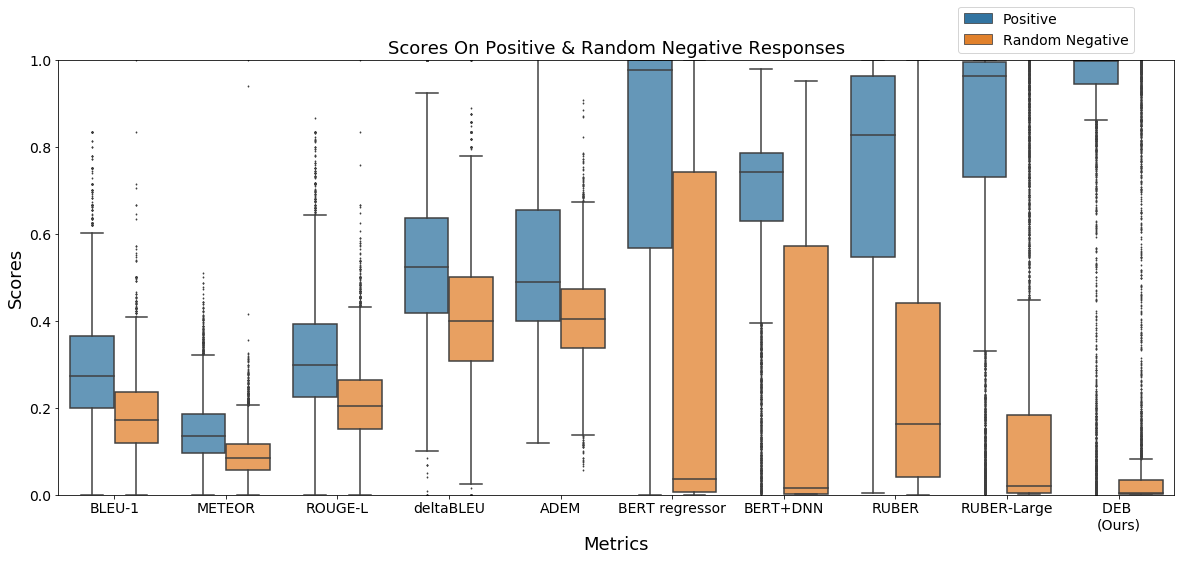}
    \caption{Box plots of the scores given by various metrics to the positive and  random negative responses}
    \label{fig:eval_rand_negs}
\end{figure*}
\subsubsection{Analysis using Box Plots \label{sec:box_plots_scores}}
We now visualize the box plots of the scores given by the various metrics to the positive and random negative responses. Figure \ref{fig:eval_rand_negs} shows these box plots for the multi-reference untrained metrics (max aggregation) and the trained metrics. 
We observe several shortcomings of the untrained metrics. Firstly, all the untrained metrics have a significant overlap in the interquartile range of the positive and random negative scores, implying that there is a high degree of intermixing of scores given to the positive and random negative responses. The overlap is even higher for word embedding based metrics, which obtain low point biserial correlations. Secondly, we note that the score distributions of the untrained metrics are highly skewed. For instance, the scores of BERTScore are almost always greater than 0.75 even though it scores responses in the range [0,1]. Therefore, it is difficult to tell at what value of the metric a response can be safely considered relevant. These observations suggest that untrained metrics even with multiple references cannot be reliably used to score dialogue responses. 

For the ADEM evaluation model, we observe that it outputs scores close to the mean score of 0.5 with little spread in their values. \newcite{DBLP:conf/aaai/SaiGKS19Reeval} also made similar observation about the clustering of the scores around the mean in ADEM, which they explain using linear system theory. In BERT regressor, there is a high overlap in the scores given to positives and random negatives. We further observe that the RUBER and BERT+DNN are able to better distinguish the positive and random negative responses. Although there is separation in the interquartile range for the two classes in RUBER and BERT+DNN scores, there is a greater spread within each class and a lot of points of the two classes substantially overlap. RUBER-Large is able to reduce the overlap, while DEB further achieves better performance by pushing the scores for positive responses close to 1 and the scores for random negatives to 0 with high accuracy. We shall show in Section \ref{sec:conicity} that DEB achieves this by pushing the $\mathbf{H}_{cls}$ embeddings for the positive and random negative responses farther apart in space.

\subsection{Performance on synthetically crafted adversarial responses}
\newcolumntype{L}[1]{>{\raggedright\let\newline\\\arraybackslash\hspace{0pt}}m{#1}}
\newcolumntype{C}[1]{>{\centering\let\newline\\\arraybackslash\hspace{0pt}}m{#1}}
\newcolumntype{R}[1]{>{\raggedleft\let\newline\\\arraybackslash\hspace{0pt}}m{#1}}

\begin{table}[t]
\centering
\resizebox{1.0\columnwidth}{!}{
\begin{tabular}{l|L{1.0cm}|L{1.28cm}|L{1.28cm}|L{1.3cm}}
\hline
Modification                                                      & DEB    & RUBER-Large  &  RUBER & BERT+DNN \\ \hline
                                                                  & \multicolumn{3}{c}{\% classified as positive}                                      \\ \hline
Unmodified positives                                              & 87.9\% &  81.7\%  & 77.5\% &  93.5\%                             \\ 
Reverse word order                                                & 60.0\% &  70.3\%  & 71.3\% & 80.4\%                             \\
Jumble word order                                                 & 69.3\% &  71.2\%  & 72.3\% & 77.4\%                             \\ 
Retain only nouns                                                 & 60.1\% &  27.9\%  & 27.8\% & 0.0\%                              \\ 
Remove punctuation                                                & 86.4\% &  72.9\%  & 72.4\% & 88.5\%                             \\ 
Remove stopwords                                                  & 85.8\% &  73.6\%  & 69.6\% & 29.3\%                             \\ 
Replace with synonyms                                             & 81.2\% &  70.8\%  & 65.6\% & 91.1\%                  \\ \hline
                                                                  & \multicolumn{3}{c}{Pearson Correlation with human scores}              \\ \hline
Remove stopwords                                                  & 0.58 \fontsize{9}{9}\selectfont{(<1e-9)} & 0.56 \fontsize{9}{9}\selectfont{(<1e-9)}  &     0.52 \fontsize{9}{9}\selectfont{(<1e-9)} & 0.056 \fontsize{9}{9}\selectfont{(0.26)}                              \\ 
\begin{tabular}[c]{@{}l@{}}Replace with synonyms\\ \end{tabular} & 0.68 \fontsize{9}{9}\selectfont{(<1e-9)} & 0.57 \fontsize{9}{9}\selectfont{(<1e-9)} &  0.54 \fontsize{9}{9}\selectfont{(<1e-9)} & -0.017 \fontsize{9}{9}\selectfont{(0.67)}                          \\ \hline 
\end{tabular}}
\caption{Fraction of responses classified as positives with synthetic modifications. Unmodified positives are presented in the 1st row for reference (p-values for individual correlations in brackets)}
\label{tab:attacks}
\end{table}
         
         
         
         
Due to space constraints, in the remainder of this section we present results only for the best performing evaluation metrics from Table \ref{table:eval_random}, viz., BERT+DNN, RUBER, RUBER-Large and DEB. Before evaluating them using the adversarial examples in our dataset, we first investigate the performance of the models with synthetically crafted adversarial attacks, similar to \newcite{DBLP:conf/aaai/SaiGKS19Reeval}. In particular, we perform simple transformations on relevant responses by (i) jumbling words in the sequence, (ii) reversing the sequence, (iii) dropping all words except nouns, (iv) dropping all stop words, (v) dropping punctuation and (vi) replacing words with synonyms.  These results are presented in Table \ref{tab:attacks}.

The modifications of reversing and jumbling the word order in a relevant response make it irrelevant (grammatically wrong) and hence we expect to see more of the original true positives get classified as negatives. BERT+DNN classifies a majority of these responses as positives. One possible reason for this is that their model only uses a max pooled aggregation on BERT embeddings and does not explicitly model the sequential order of words. On the other hand, DEB fares better than the other models as seen by the drop in fraction of responses identified as positives. However, RUBER variants and BERT+DNN do better than DEB when retaining only nouns in a response. \Ananya{On removing punctuation, we expect that most of the positive responses without punctuation would remain positive and hence the percentage of responses marked positive should remain about the same. In this case, both DEB and BERT+DNN perform better than the RUBER models.} For the modifications of removing stopwords and replacing words with synonyms, it is hard to generalize the trend that is observed. Hence, we perform human evaluations by presenting in-house annotators with contexts and modified responses.  We ask them to provide scores in the range 0 to 3, with higher scores meaning better responses. We obtain human scores on 400 samples for this task and compute the Pearson correlation of the model predictions with the human judgements. In this case, we find DEB is better correlated with human judgements on both the modifications. 

\begin{figure}
    \centering
    \includegraphics[width=1.0\columnwidth]{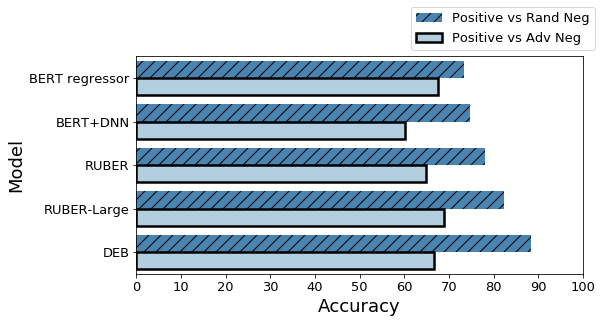}
    \caption{Accuracy of different models in identifying adversarial and random negatives versus positive responses}
    \label{fig:rand_vs_adv_negs}
\end{figure}

\subsection{Performance of model-based metrics on manually crafted adversarial responses}
         
         
         
         
        
\begin{table}[t]
    \centering
    \resizebox{1.0\columnwidth}{!}{
    \begin{tabular}{C{2.1cm}C{2.1cm}C{2.1cm}}
         \colorbox{springbud}{TP} \colorbox{palerobineggblue}{FN}
 \colorbox{yellow!70}{FP} \colorbox{wildwatermelon}{TN}  & Positive vs Random negatives & Positive vs Adversarial negatives \\

         BERT+DNN & \colorbox{springbud}{5337} \colorbox{palerobineggblue!70}{ 373 }
 \colorbox{pastelyellow!70}{2520} \colorbox{wildwatermelon}{3190} & \colorbox{springbud}{5337} \colorbox{palerobineggblue!70}{ 373 }
 \colorbox{yellow!40}{4179} \colorbox{wildwatermelon!60}{1531}\\
        & & \\
        
        BERT regressor & \colorbox{springbud}{3442} \colorbox{palerobineggblue!70}{1126}
 \colorbox{pastelyellow!70}{1304} \colorbox{wildwatermelon}{3264} & \colorbox{springbud}{3442} \colorbox{palerobineggblue!70}{1126}
 \colorbox{yellow!40}{1837} \colorbox{wildwatermelon!60}{2731}\\
        & & \\

         RUBER & \colorbox{springbud}{4420} \colorbox{palerobineggblue!70}{1280}
 \colorbox{pastelyellow!70}{1207} \colorbox{wildwatermelon}{4493} & \colorbox{springbud}{4420} \colorbox{palerobineggblue!70}{1280}
 \colorbox{yellow!40}{2714} \colorbox{wildwatermelon!60}{2986}\\
        & & \\
        
        RUBER-Large & \colorbox{springbud}{4659} \colorbox{palerobineggblue!70}{1041}
 \colorbox{pastelyellow!70}{970} \colorbox{wildwatermelon}{4730} & \colorbox{springbud}{4659} \colorbox{palerobineggblue!70}{1041}
 \colorbox{pastelyellow!70}{2500} \colorbox{wildwatermelon}{3200}\\
        & & \\
        
        DEB & \colorbox{springbud}{5011} \colorbox{palerobineggblue!70}{\ 689\ }
 \colorbox{pastelyellow!70}{\ 646\ } \colorbox{wildwatermelon}{5054} & \colorbox{springbud}{5011} \colorbox{palerobineggblue!70}{\ 689\ }
 \colorbox{yellow!60}{3101} \colorbox{wildwatermelon!60}{2599} \\
    \end{tabular}
    }
    \caption{Confusion matrix showing changes in the performance of different models on DailyDialog++ with random and adversarial negatives.}
    \label{tab:bert_conf}
\end{table}
So far we have established that (i) untrained metrics perform poorly compared to trained metrics even for separating random negatives from positives (ii) trained models like RUBER, BERT+DNN, RUBER-Large and DEB perform remarkably well in distinguishing relevant responses from random responses (iii) RUBER variants and DEB perform well on most synthetically mutated responses whereas BERT+DNN performs poorly against certain mutations. However, we still need to check if the trained models are robust to adversarial examples which are specifically crafted to fool such \textit{context-dependent}, model-based metrics. Note that none of the untrained metrics are context dependent as they directly compute the similarity between the reference and candidate response without considering the context.

We consider the 5 relevant and the 5 adversarial irrelevant responses in our dataset and just as before compute the scores assigned by the different metrics to each of these responses. We then compute the accuracy of a metric using the target label as 0 for irrelevant responses and 1 for relevant responses.
As expected, the accuracy of all the models drops, as seen in Figure \ref{fig:rand_vs_adv_negs}. 
In particular, we observe that the models wrongly classify most of the irrelevant responses as positive/relevant responses. This can be seen from the confusion matrices in Table \ref{tab:bert_conf}, where it is clear that the number of false positives is very high. 
\section{Discussions}
In this section, we do further analysis of DEB.
\subsection{Ablation studies on DEB \label{sec:ablation}}
There are different stages of training our DEB model. First, the underlying BERT model is already pretrained on English Wikipedia and the BooksCorpus. We then pretrain it further for our task using Reddit corpus and finally finetune it on the DailyDialog++ dataset. We now evaluate the contributions of each of these stages of training (see Table \ref{tab:abl}). 
First, we find that the original BERT model when adopted directly for the task of dialog evaluation gives an accuracy of 72.65\% and 58.10\% on random and adversarial negatives respectively. On further analysis, we find that it has a high false positive rate with more than 52\% of the adversarial negatives getting classified as positives. 
After pretraining it with Reddit data, it achieves an accuracy of 84.16\% on DailyDialog++ even though it has not seen any training instances from this dataset. However, there is only a marginal improvement on adversarial negatives. 
Finally, finetuning BERT on DailyDialog++ using only random negatives further improves the accuracy to 88.29\% and 66.75\% respectively. 
    \begin{table}[t]
        \centering
        \resizebox{1.0\columnwidth}{!}{
        \begin{tabular}{|c|C{1.54cm}|C{1.46cm}|}
        \hline
         Model & Pos vs Rand Neg & Pos vs Adv Neg\\
         \hline
         BERT original & 72.65  & 58.10 \\ 
         DEB pretrained on Reddit & 84.16 & 59.82 \\
         Pretrained DEB finetuned on rand neg & 88.29 & 66.75 \\
             \hline
        \end{tabular}
        }
        \caption{Ablation studies on DEB}
        \label{tab:abl}
    \end{table}
    \begin{table}[t]
        \centering
        \resizebox{1.0\columnwidth}{!}{
        \begin{tabular}{|C{2.48cm}|C{3.0cm}|C{1.54cm}|C{1.45cm}|}
        \hline
         Model & Training/ Finetuning Data & Pos vs Rand Neg & Pos vs Adv Neg\\
         \hline
         \multirow{3}{*}{BERT regressor} & Rand neg & 73.40  & 67.57 \\ 
          & Adv neg & 69.89  & 75.92 \\ 
          & Rand + Adv neg & 72.77 & 74.55 \\ 
         \hline
         \multirow{3}{*}{BERT+DNN}  & Rand neg & 74.67 & 60.14 \\
          & Adv neg &  60.49 & 87.67 \\
          & Rand + Adv neg & 73.87 & 86.61\\ 
         \hline
         \multirow{2}{*}{RUBER} & Rand neg & 78.18  & 64.96 \\ 
          & Adv neg & 70.82  & 76.50 \\ 
          \fontsize{10}{10}\selectfont{ (Pretrained)} & Rand + Adv neg & 75.11  & 83.88 \\ 
         \hline
         \multirow{2}{*}{RUBER-Large } & Rand neg & 82.35  & 68.94 \\ 
          & Adv neg & 63.99  & 90.49 \\ 
          \fontsize{10}{10}\selectfont{ (Pretrained)} & Rand + Adv neg & 79.91  & 86.54 \\ 
         \hline
          \multirow{2}{*}{DEB } & Rand neg & 88.29 & 66.75 \\
          & Adv neg & 86.24 & 82.04 \\
           \fontsize{10}{10}\selectfont{ (Pretrained)} & Rand + Adv neg &\textbf{88.67} &\textbf{92.65} \\ \hline
        \end{tabular}
        }
        \caption{Accuracy in classifying Pos vs Rand Neg and Pos vs Adv Neg responses for various model variants trained/finetuned on DailyDialog++.}
        \label{tab:accuracies}
    \end{table}
\subsection{Training with adversarial examples}
\begin{figure}
    \centering
    \includegraphics[width=1.0\columnwidth]{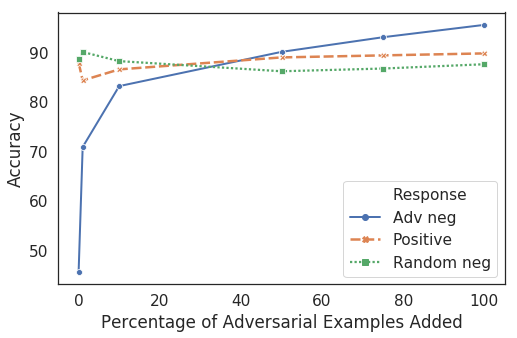}
    \caption{Effect of varying the amount of adversarial negatives added to the training set}
    \label{fig:adv_size}
\end{figure}

We examine whether the evaluation models can learn to distinguish the adversarial negatives when specifically finetuned for that task. By training on DailyDialog++ with adversarial negatives rather than random negatives, we find that all models give an improved performance in identifying adversarial negatives (see Table \ref{tab:accuracies}). 
However, with such training, every model's performance drops when evaluated on DailyDialog++ with random negatives, with BERT+DNN dropping substantially to 60.49\%. 
The best overall performance is seen when the models are finetuned with both random and adversarial negatives, with DEB achieving the highest accuracies on both test sets. While such improvement is expected given the capacity of the models, obtaining such adversarial examples for training is not always feasible.\\
\noindent\textbf{Effect of the number of adversarial negatives added to training:}
Due to the difficulty in manually creating adversarial examples, we study the effect of the number of the adversarial examples added to the training set. Our findings are presented in Figure \ref{fig:adv_size}, where we progressively increase the percentage of adversarial negative examples added as input to the DEB model during training with random negatives. 
As expected, the accuracy in identifying adversarial negatives improves as the model is exposed to more data points of the same type, where we specifically note the considerable improvement from 45.6\% to 70.85\% after adding just 1\% of adversarial negatives from our dataset (\textit{i.e.}, 100 contexts with 5 adversarial examples each). With the addition of more adversarial negatives, we find a small drop in the accuracy of identifying random negatives. There is also a slight decrease in the performance on the positives responses when the number of adversarial examples are small. We note that the adversarial negatives are hard negatives close to the positive responses in the embedding space, as we elaborate in Section \ref{sec:conicity}, thereby confusing the model.

\subsection{Conicity analysis on DEB \label{sec:conicity}}
\begin{figure*}
\centering
\begin{subfigure}{0.5\textwidth}
  \centering
  \includegraphics[width=.5\textwidth]{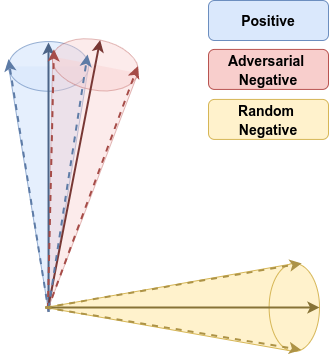}
  \caption{Before finetuning on DailyDialog++}
  \label{fig:sub1}
\end{subfigure}%
\begin{subfigure}{.5\textwidth}
  \centering
  \includegraphics[width=0.75\textwidth]{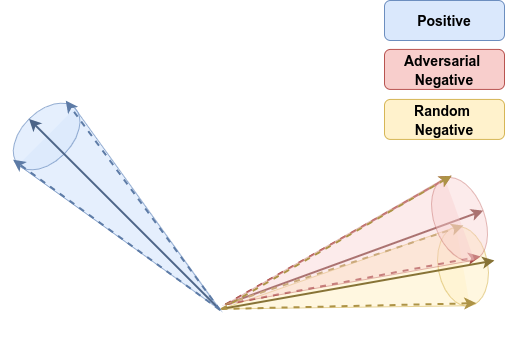}
  \caption{After finetuning on DailyDialog++}
  \label{fig:sub2}
\end{subfigure}
\caption{Illustration of the spread of the positive and negative response embeddings by DEB \small{(not to scale)}}
\label{fig:conicity}
\end{figure*}
We analyze the embeddings from the final embeddings projection space, that is, the one used by softmax layer for next response prediction. We check for the spread of the embeddings of the positive and negative responses. Specifically, let $P,R$ and $A$ be the set of embeddings of all positive responses, random negative responses and adversarial negative responses respectively for a given context. We want that if we consider the set $P$ then the spread of this set should be low in the projected space (all positive responses embedded close to each other). At the same time, if we consider the union of the sets $P, R$ and $A$ then the spread of this set should be high (positive responses separated from negative responses). We measure this spread using conicity analysis \cite{DBLP:conf/acl/TalukdarSC18conicity}. Conicity on a set of vectors $V$ is defined as the average of the cosine similarity of the vectors with their mean vector, $\bar{v}$.
The lower the conicity, the higher the spread.

For each utterance in DailyDialog++, we first construct the sets $P,R$ and $A$ using the pretrained DEB model. We find that the average conicity of the set $P$ is 0.89 (averaged over all utterances) indicating that the positive responses get mapped very close to each other. The average conicity of the set $P \cup R$ is 0.59, indicating that the positive responses are well separated from the random negatives. However, the average conicity of the set $P \cup A$ is 0.74, indicating that the positive responses are not well separated from the adversarial negative responses. We illustrate this in Figure \ref{fig:sub1} \Ananya{by representing the mean vector of each of the sets along a corresponding highlighted region where the vectors of the set lie on average}.\footnote{Note that separation of cones in the figure does not indicate complete separation of all the vectors between the sets, rather separation on average, as there could be some overlap or outliers, as evident from the model's performance in various experiments.}
We then finetune the DEB model on the DailyDialog++ dataset. Once again, for every utterance we construct the sets $P,R$ and $A$ using this finetuned model. We now observe that the average conicity of the sets $P$, $P \cup R$ and $P \cup A$ are 0.86, 0.37 and 0.35 respectively. Thus, after finetuning, the model is able to achieve a clear separation between positive responses and random or adversarial negative responses. Furthermore, the positive responses are still close to each other (illustrated in Figure \ref{fig:sub2}).

\section{Generalization to other datasets}

In this section, we investigate how well the different model-based metrics trained on DailyDialog++ generalize to other datasets which are not seen during training. We evaluate the 3 unreferenced models, BERT+DNN, RUBER, and DEB, which require only context and candidate response as inputs on these 3 datasets.\\
\textbf{Twitter:} Microsoft Research Social Media Conversation Corpus \cite{DBLP:conf/naacl/SordoniGABJMNGD15} contains a curated list of 3-turn Twitter conversations, all of which are human-verified as good responses.\\
\textbf{PersonaChat:} The dialogues in PersonaChat \cite{personachat} are associated with well-defined personalities of the speakers involved. 
We consider the verified human-human chat logs, released by \newcite{DBLP:conf/naacl/SeeRKW19}, as positive examples.\\
\textbf{Holl-E:} This dataset \cite{DBLP:conf/emnlp/MogheABK18Holle} contains conversations about movies, where each response is generated by copying and modifying content from a relevant background document. 
We use the multi-reference test set of Holl-E containing 4 positive responses for each context. 
\\
For all the 3 datasets, we consider the reference responses as positive responses and obtain negative examples by randomly sampling responses from other contexts. We reiterate that we do not train the models on these datasets but simply evaluate the models trained on DailyDialog++ on these datasets. Table \ref{tab:datasets} shows that DEB outperforms the other unreferenced models on all the 3 datasets. With Holl-E dataset being specific to conversations about movies rather than generic topics, we find the scores are relatively lower on it for all the models. The other evaluation models and metrics cannot be compared on PersonaChat and Twitter without additional reference responses, since the available single reference in these datasets is being evaluated. 
On the multi-reference test set of Holl-E, however, we find that their performance is lower than the three unreferenced models.
\newcolumntype{L}[1]{>{\raggedright\let\newline\\\arraybackslash\hspace{0pt}}m{#1}}
\newcolumntype{C}[1]{>{\centering\let\newline\\\arraybackslash\hspace{0pt}}m{#1}}
\newcolumntype{R}[1]{>{\raggedleft\let\newline\\\arraybackslash\hspace{0pt}}m{#1}}

\begin{table}[t]
\centering
\resizebox{1.0\columnwidth}{!}{
\begin{tabular}{|L{3.02cm}|l|l|l|}
\hline
        Model  & Persona & Twitter & Holl-E \\\hline
         BERT+DNN & 71.01 & 48.71 & 54.60\\
         RUBER  & 61.17 & 71.18 & 54.83\\
         RUBER-Large & 62.32 & 77.18 & 55.94 \\
         DEB & \textbf{78.55} & \textbf{82.71} & \textbf{62.74}\\\hline
   \end{tabular}}
\caption{Transferability to other datasets}
\label{tab:datasets}
\end{table}
\section{Correlations with human judgements on system generated responses}
\MK{Lastly, we wanted to check if DEB scores correlate well with scores assigned by humans on responses generated by dialogue systems (as opposed to humans). To do so},
we collected responses generated by the following five dialogue response generation models: 


\noindent \textbf{HRED:} Hierarchical Recurrent Encoder Decoder (HRED) \cite{HRED} extends the traditional seq2seq model by adding an additional utterance-level RNN. 

\noindent \textbf{VHRED:} Latent Variable HRED (VHRED) \cite{VHRED} includes a latent variable
at the decoder, and is trained by maximizing a variational lower-bound on the log-likelihood. 

\noindent \textbf{VHCR:}  Variational Hierarchical Conversation RNN (VHCR) \cite{VHCR} further extends VHRED by drawing a prior encoding for each conversation. 

\noindent \textbf{DialoGPT small:}  \newcite{zhang2019dialogpt} pretrained GPT-2-like \cite{gpt2} transformer models on 147M conversations extracted from Reddit comments. The small version contains 12 layers and 768 hidden dimensions. 

\noindent \textbf{DialoGPT medium:} The medium version of DialogGPT contains 24 layers and 1024 hidden dimensions.

For the RNN-based models (HRED, VHRED, VHCR), we use a single-layer bidirectional encoder and single-layer decoder each with a hidden size of 1024. We pretrain the RNN-based models on the casual conversation subset of the Reddit dataset, consisting of 10M conversation exchanges. We finetune all the models on the DailyDialog++ dataset. 

We conducted human evaluations to compare the extent to which the model-based metrics agree with human judgements. We randomly sampled 100 contexts from the test set of the DailyDialog++ dataset and obtained the responses generated by each of the above models. Annotators were shown a context-response pair and were asked to rate how human-like the response is with respect to the context, on a scale of 0-3. The annotators were asked to check for both fluency and coherence. A total of 15 in-house annotators participated in the human evaluation study. The annotators were Computer Science graduates competent in English. Each context-response pair was rated by 5 annotators and the final score was obtained by averaging the 5 scores.  
We also obtained scores at the system level by aggregating the scores for each model. In Table \ref{tab:human-eval}, we report the correlations of human judgments with the model scores at the response level and system level. 
We observe that the BERT+DNN model, which only has a feed-forward neural network that is learnable, does not have any significant correlation with human judgments. On the other hand RUBER, consisting of pretrained GRUs obtains low to moderate correlations. \Ananya{RUBER-Large further obtains improved correlations, indicating that using large-scale pretrained models helps. This trend is also observed in the comparisons of DEB  with its ablated versions (without Reddit pretraining and without finetuning on DailyDialog++), indicating the contribution of these steps in training the final model. Our proposed DEB model obtains significantly higher correlations at response level. We checked for significance using William's test to compare DEB with all other models and found p-values to be $<1 e^{-6}$. This establishes the effectiveness of DEB in scoring model generated responses. At the system level, we find that DEB correlates substantially higher than other models, with the human rankings of the models. However, the p-values in this case are not significant due to the limited number of systems. \MK{In hindsight, we realise that reporting system level correlations is not very informative as the number of samples are very small (as many as the number of systems). Hence, these numbers are not very reliable. However, following \newcite{DBLP:conf/acl/LoweNSABP17ADEM}, we still report the system-level correlations (along with the p-values) for the sake of completeness.}} 

\newcolumntype{L}[1]{>{\raggedright\let\newline\\\arraybackslash\hspace{0pt}}m{#1}}
\newcolumntype{C}[1]{>{\centering\let\newline\\\arraybackslash\hspace{0pt}}m{#1}}
\newcolumntype{R}[1]{>{\raggedleft\let\newline\\\arraybackslash\hspace{0pt}}m{#1}}

\begin{table}[]
\resizebox{1.0\columnwidth}{!}{
\begin{tabular}{l|l|l|l}
\hline

\multicolumn{1}{c|}{Model} & \multicolumn{1}{l|}{Pearson} & \multicolumn{1}{l|}{Spearman} & \multicolumn{1}{l}{Kendall tau} \\ \hline
\multicolumn{4}{c}{{Response level}}                                                                                          \\ \hline
BERT+DNN                    & 0.016 \fontsize{9}{9}\selectfont{(0.73)}                       & 0.009 \fontsize{9}{9}\selectfont{(0.89)}                        & 0.007 \fontsize{9}{9}\selectfont{(0.88)}                            \\ 
RUBER                       & 0.111 \fontsize{9}{9}\selectfont{(2.5e-2)}                       & 0.126 \fontsize{9}{9}\selectfont{(1.1e-2)}                        & 0.090 \fontsize{9}{9}\selectfont{(8.9e-2)}  \\

RUBER-Large                 & 0.265 \fontsize{9}{9}\selectfont{(<1e-7)}                       & 0.256 \fontsize{9}{9}\selectfont{(<1e-6)}                        & 0.173 \fontsize{9}{9}\selectfont{(<1e-6)}  \\ \hline
DEB w/o Reddit & 0.356  \fontsize{9}{9}\selectfont{(<1e-9)}                   & 0.295 \fontsize{9}{9}\selectfont{(<1e-9)}                        & 0.202 \fontsize{9}{9}\selectfont{(<1e-9)}\\
DEB w/o DD++ & 0.274  \fontsize{9}{9}\selectfont{(<1e-9)}                   & 0.337 \fontsize{9}{9}\selectfont{(<1e-9)}                        & 0.232 \fontsize{9}{9}\selectfont{(<1e-9)}\\
DEB                         & \textbf{0.440*}  \fontsize{9}{9}\selectfont{(<1e-9)}                       & \textbf{0.523*} \fontsize{9}{9}\selectfont{(<1e-9)}                        & \textbf{0.374*} \fontsize{9}{9}\selectfont{(<1e-9)}                            \\ \hline
\multicolumn{4}{c}{{System level}}                                                                                            \\ \hline
BERT+DNN                   & 0.050   \fontsize{9}{9}\selectfont{(0.89)}                     & -0.100 \fontsize{9}{9}\selectfont{(0.87)}                       & 0.000 \fontsize{9}{9}\selectfont{(1.1) }                            \\ 
RUBER                       & 0.221 \fontsize{9}{9}\selectfont{(0.72)}                      & 0.300 \fontsize{9}{9}\selectfont{(0.62)}                        & 0.200 \fontsize{9}{9}\selectfont{(0.81)} \\
RUBER-Large                 & 0.679 \fontsize{9}{9}\selectfont{(0.20)}                      & 0.499 \fontsize{9}{9}\selectfont{(0.39)}                        & 0.399 \fontsize{9}{9}\selectfont{(0.483)} \\ \hline
DEB w/o Reddit & 0.784   \fontsize{9}{9}\selectfont{(0.12)}                    & 0.600 \fontsize{9}{9}\selectfont{(0.28)}                         & 0.400 \fontsize{9}{9}\selectfont{(0.48)}\\
DEB w/o DD++ & 0.855  \fontsize{9}{9}\selectfont{(0.06)}                   & 0.600 \fontsize{9}{9}\selectfont{(0.28)}                         & 0.400 \fontsize{9}{9}\selectfont{(0.48)}\\
DEB                         & \textbf{0.973}    \fontsize{9}{9}\selectfont{(5.2e-3)}                    & \textbf{0.700} \fontsize{9}{9}\selectfont{(0.18)}                         & \textbf{0.600} \fontsize{9}{9}\selectfont{(0.23)}                            \\ \hline
\end{tabular}}
\caption{Human correlations on DailyDialog++ data with different models. (Individual p-values in parenthesis.) * indicates statistical significance in performance over other models, with p-values <1e-6 on the William’s test} 
\label{tab:human-eval}
\end{table}

\section{Related Work}

We point the reader to \newcite{DBLP:journals/corr/SerbanLCP15} for an excellent survey of existing datasets containing single reference responses. Recently, there has been some effort to create datasets containing multiple references but these datasets are either too small (around 1000 contexts) \cite{DBLP:conf/emnlp/MogheABK18Holle, DBLP:journals/corr/abs-1907-10568InvestEval} or noisy \cite{DBLP:conf/naacl/GaoLZBGGD19redditmultiforgenerationdiversity}. 

We have already reviewed all the existing dialog metrics in Section \ref{sec:existing-metrics} and hence we do not discuss them again here. Instead, we quickly mention existing works which critically examine dialog evaluation metrics. For example, \newcite{DBLP:conf/emnlp/LiuLSNCP16HowNot} show that existing $n$-gram based metrics do not correlate well with human judgements for dialog evaluation. We report similar results but additionally show that the correlation improves in the presence of multiple references. 
Similarly, \newcite{DBLP:conf/aaai/SaiGKS19Reeval} have critically examined ADEM and shown that in most cases it produces a score close to 2.5 (on a scale of 1 to 5) and hence does not clearly separate relevant and irrelevant responses.

Lastly, we also mention a very recent work, \newcite{zhang2019dialogpt}, which has pretrained a large scale transformer on Reddit corpus for building conversation systems. However, their focus is on dialog generation and not on evaluation metrics. 


\section{Conclusions}
We propose a multi-reference open-domain dialogue dataset with multiple relevant responses and adversarial irrelevant responses. We perform an extensive study of the existing dialogue evaluation metrics using this dataset and also propose a new transformer-based evaluator pretrained on large-scale dialogue datasets. We identify the strengths and weaknesses of such a model through studies of its performance on untrained and synthetically modified data. We find DEB to be easily adaptable to other open-domain dialogue datasets. We also present the scope of the adversarial responses in our dataset towards bringing out better evaluation metrics, since all the current models do not perform well on those unless explicitly trained. 

\iftaclpubformat

\section*{Acknowledgments}
We thank the Department of Computer Science and Engineering, IIT Madras and the Robert Bosch Center for Data Science and Artificial Intelligence, IIT Madras (RBC-DSAI) for providing us resources required to carry out this research. We are grateful to Google for the TFRC credits that supported our usage of TPUs for several experiments in this paper. We also thank Google for supporting Ananya Sai through their Google India Ph.D. Fellowship Program. 
We thank the action editor, Xiaojun Wan, and all the anonymous reviewers for their very helpful comments in enhancing the work. We thank the in-house human annotators and evaluators for helping us create the dataset.
\else
\fi

         

\bibliography{tacl2018}
\bibliographystyle{acl_natbib}

\end{document}